\pdfoutput=1

\documentclass[11pt]{article}

\usepackage[]{acl}
 
\usepackage{times}
\usepackage{latexsym}

\usepackage[T1]{fontenc}

\usepackage[utf8]{inputenc}

\usepackage{microtype}

\usepackage{graphicx}
\usepackage{enumitem}
\usepackage{multirow}
\usepackage{adjustbox}
\usepackage{booktabs}

\title{Predicting the Quality of Revisions in Argumentative Writing}

\author{Zhexiong Liu$^{1}$, Diane Litman$^{1,2}$, Elaine Wang$^{3}$, Lindsay Matsumura$^{2}$, Richard Correnti$^{2}$
 \\
    $^{1}$Department of Computer Science \\
  $^{2}$Learning Research and Development Center \\
University of Pittsburgh, Pittsburgh, Pennsylvania 15260 USA\\
$^{3}$RAND Corporation, Pittsburgh, Pennsylvania 15213 USA \\
   \texttt{zhexiong@cs.pitt.edu}, \texttt{ewang@rand.org} \\ \texttt{\{dlitman,lclare,rcorrent\}@pitt.edu}
  }

\begin{document}
\maketitle
\begin{abstract}

The ability to revise in response to feedback is critical to students' writing success. In the case of argument writing in specific, identifying whether an argument revision (AR) is successful or not is a complex problem because AR quality is dependent on the overall content of an argument. For example, adding the same evidence sentence could strengthen or weaken existing claims in different argument contexts (ACs). To address this issue we developed Chain-of-Thought prompts to facilitate ChatGPT-generated ACs for AR quality predictions. The experiments on two corpora, our annotated \textit{elementary essays} and existing \textit{college essays} benchmark, demonstrate the superiority of the proposed ACs over baselines.

\end{abstract}

\section{Introduction} 
\begin{figure}[t]
    \centering
    \includegraphics[width=0.95\columnwidth]{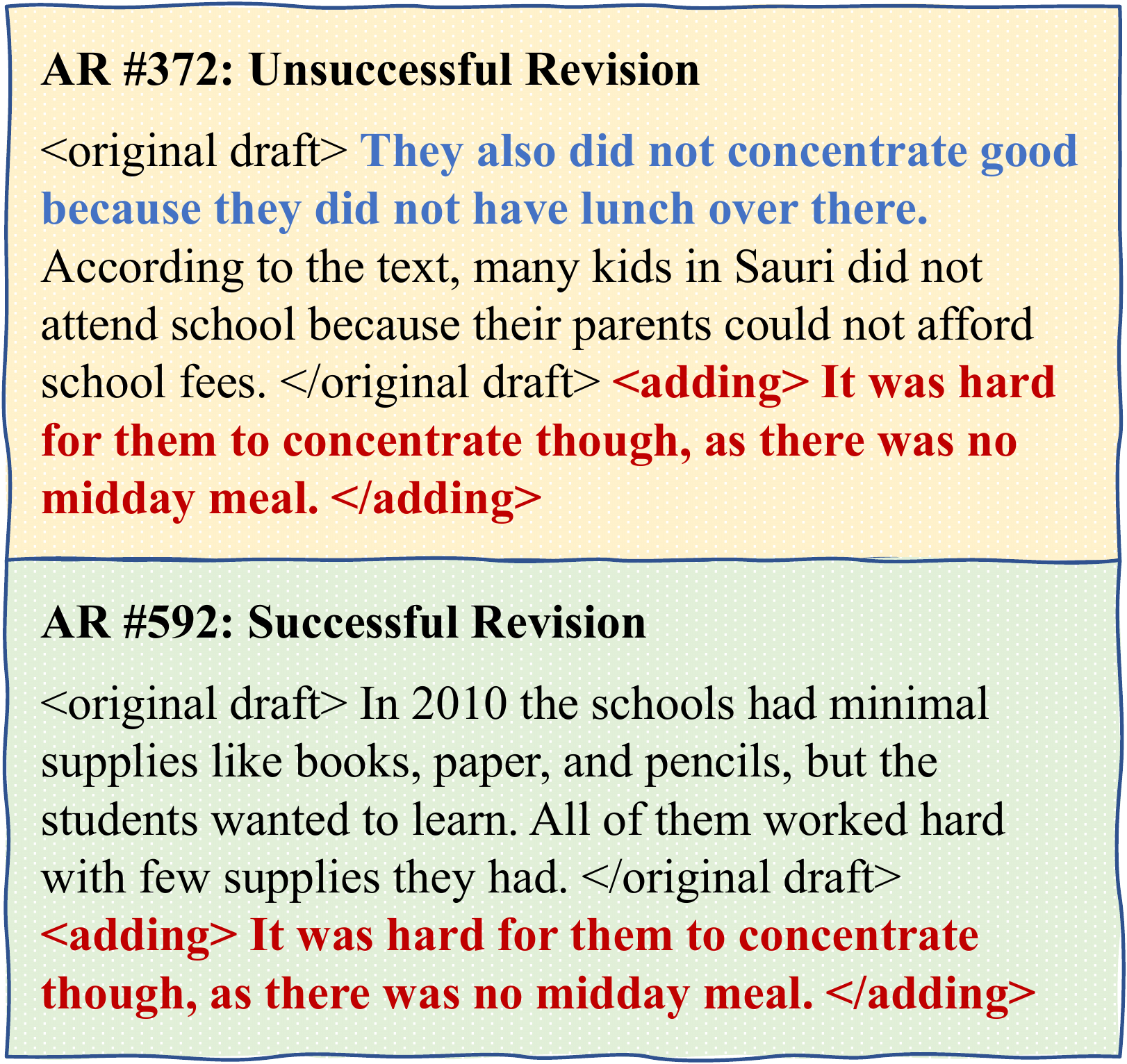}
    \caption{Two pieces of ARs in two student essays show that \textit{adding} the same sentence \textit{``it was hard for them to concentrate though, as there was no midday meal''} (bold in red) in different contexts caused opposite AR quality. AR \#372 added a piece of evidence that \textit{already existed} in the original draft (bold in blue) thus the attempted AR did not improve the essay quality. AR \#592 improved the quality by adding a \textit{relevant} piece of new evidence. AR \#372 was \textit{unsuccessful} while AR \#592 was \textit{successful}.}
    \label{fig:example}
\end{figure}

Argumentative Revision (AR) in response to feedback is important for improving the quality of students' written work. Successful ARs\footnote{\citet{afrin2023predicting} use the term desirable revisions.} usually include adding relevant evidence, deleting repeated evidence or reasoning, and elaborating relevant evidence examples to support claims~\cite{afrin2020RER}. Differentiating between successful versus unsuccessful ARs, however, is a complex endeavor. For example, making the same AR in distinct Argumentative Contexts (ACs) could differentially affect the quality of a student's essay. Here the ACs are defined as pieces of sentences that present reasons, evidence, and claims supporting or opposing arguments in argumentative writing (see Sec.~\ref{sec:context}). For example, Figure \ref{fig:example} shows two pieces of ARs that added the same sentence \textit{``it was hard for them to concentrate though, as there was no midday meal''} but caused opposite AR quality.

Recently developed Automated Writing Evaluation (AWE) systems have focused on assessing the content and structure of student essays to automatically provide students with formative feedback~\cite{zhang2016argrewrite,ets-writing-mentor,wang2020eRevise,beigman2020AWE}. Successful revisions (e.g., adding relevant evidence) improve an essay’s quality. Unsuccessful revisions, in contrast, lead to no improvement or can even weaken an essay’s argument~\cite{afrin2020RER}. As a result, assessing the success of ARs is important to assess the quality of ARs in line with provided feedback. 

AR quality has previously been predicted by using long and short neighboring contexts of ARs~\cite{afrin2023predicting}. This location-based approach for constructing ACs  did not exploit any argumentative relationships between ARs and potential ACs. Another study~\cite{zhang2016using} incorporated AR contexts with cohesion blocks and employed sequence labeling to model AR interdependence across revisions. This work predicted AR purposes from discourse structures but did not further study AR quality or analyze AR quality from the perspective of ACs. To bridge these gaps, we address three research questions. \textbf{RQ1}: To what extent are ACs  helpful for predicting AR quality? \textbf{RQ2}: What type of AC is the most helpful in AR quality predictions? \textbf{RQ3}: Can ChatGPT prompts be used to generate useful ACs? In studying the three RQs, we have made the following contributions:

\begin{itemize}[leftmargin=*,itemsep=-5pt, topsep=3pt]
  \item Our project is the first in the revision field to analyze the relationship between ACs and AR quality predictions.
  \item We are among the first to incorporate the state-of-the-art large language model ChatGPT in generating ACs in argumentative writing.
  \item Experiments using both elementary and college essay corpora show the superiority of the proposed ACs over existing location-based contexts for AR quality predictions.
\end{itemize} 

\section{Related Work}

\subsection{Argumentative Revision in NLP} 
Revision research has been conducted using multiple types of Natural Language Processing (NLP) corpora ranging from Wikipedia to argumentative essays. While argumentative writing research has analyzed argumentative roles and discourse elements in persuasive writing~\cite{stab2014identifying,song2020discourse,putra2021parsing} (e.g., by studying the stance towards some topic, backing up claims, or following argumentative and rhetorical considerations), such analyses have not typically been applied to revision research in this domain. Revision research, in contrast, has primarily focused on grammar correction, paraphrasing, semantic editing~\cite{yang2017hkh}, and analyzing revision purposes~\cite{zhang2015l,shibani2018kb,afrin2020RER,kashefi2022}. Although revision research has sometimes leveraged contextualized features during classification, the contextual features have been location-based~\cite{zhang2016using, afrin2023predicting}. We instead  extract contextual information from an essay based on argumentative essay analysis rather than on adjacency to a revision. 

\subsection{LLM in Argumentative Revision}

\begin{table*}[t]\small
\centering
\begin{adjustbox}{width=1.6\columnwidth}
\begin{tabular}{cc|ccc|ccc}
\toprule
 &  & \multicolumn{3}{c|}{\textbf{Elementary Essays}} & \multicolumn{3}{c}{\textbf{College Essays}} \\
 &  & Reasoning & Evidence & Total & Reasoning & Evidence & Total \\ \midrule
\multirow{3}{*}{Successful} & Add & 769 & 671 & 1440 & 104 & 23 & 127 \\
 & Delete & 213 & 104 & 317 & 7 & 1 & 8 \\
 & Modify & 129 & 104 & 233 & 3 & 0 & 3 \\ \midrule
\multirow{3}{*}{Unsuccessful} & Add & 360 & 491 & 851 & 87 & 2 & 89 \\
 & Delete & 102 & 147 & 249 & 6 & 0 & 6 \\
 & Modify & 74 & 103 & 177 & 0 & 0 & 0 \\ \midrule
Total & / & 1647 & 1620 & 3267 & 207 & 26 & 233 \\ \bottomrule
\end{tabular}
\end{adjustbox}
\caption{\label{table:sentence-stats}
Sentence-level AR quality annotation statistics on \textit{elementary} and \textit{college essays}.
}
\end{table*}

Large Language Models (LLMs) have scaled up model sizes from a few million to hundreds of billions of parameters. Their strong capabilities of handling multiple downstream NLP tasks have made LLMs favorable in recent research~\cite{palm}. Prior revision works, e.g., academic writing~\cite{ito-etal-2019-diamonds}, debation assessment~\cite{skitalinskaya-etal-2021-learning}, paraphrase generation~\cite{mu-lim-2022-revision}, were mostly based on Transformer~\cite{vaswani2017attention} and BERT~\cite{devlin2019BERT} models but not the cutting-edge LLMs, e.g., ChatGPT\footnote{\url{https://openai.com/blog/chatgpt}}. Pretrained LLMs have shown strong few-shot learning capabilities by way of developing prompts to guide LLMs in generating successful outputs~\cite{brown2020language,liu2023pre}. For example, Chain-of-Thought (CoT) prompts~\citep{cot_wei} enable pretrained LLMs to solve complex reasoning problems by decomposing the tasks into a series of intermediate steps.~\citet{kojima2022large,wang2022iteratively} investigated the effectiveness of CoT in multi-step reasoning, however, little work has used CoT for extracting and then generating tasks in the revision field. In this work, we leverage ChatGPT with CoT prompts to generate ACs in argumentative writing.

\section{Corpora}

\begin{table*}[!htb]\small
\centering
\begin{adjustbox}{width=2.08\columnwidth}
\begin{tabular}{p{0.01\textwidth}p{0.26\textwidth}p{0.26\textwidth}p{0.05\textwidth}p{0.06\textwidth}p{0.08\textwidth}p{0.09\textwidth}}		
\toprule
\textbf{ID} & \textbf{Original Draft Sentence} & \textbf{Revised Draft Sentence} & \textbf{Revision} & \textbf{Revision Type} & \textbf{Revision Purpose} & \textbf{Quality Label} \\ \midrule
1 & According to the text. "The people in Sauri have made amazing progress in just eight years." & According to the text, "The people in Sauri have made amazing progress in just eight years." & Modify & Surface & N/A & N/A \\ \midrule
2 & This tells me that the people of Sauri have made better living arrangements in eight years and it all did pay off. & This tells me that the people of Sauri have made better living arrangements in eight years and it all did pay off, from how they used to live. & Modify & Content & N/A & N/A \\ \midrule
3 & The people of Sauri did do great progress. & The people of Sauri did do great progress. & N/A & N/A & N/A & N/A \\ \midrule
4 & ... & ... & ... & ... & ... & ... \\ \midrule
5 & This lets me know that since there might be a few diseases that might affect anyone at ant time so the hospital has made medicine that can cure those diseases, so they gave that medicine to any one who needed it for free. &  & Delete & Content & Irrelevant Evidence & Unsuccessful \\ \midrule
6 &  & This piece of text lets me know that the hospital, Yala Sub District, has free medicine for diseases that are most common around where they live. & Add & Content & Paraphrase Reasoning & Successful \\ \midrule
7 &  & In Sauri people had to pay a fee, which the people of Sauri couldn't afford. & Add & Content & not LCE Reasoning & Unsuccessful \\ \midrule
8 & ... & ... & ... & ... & ... & ... \\ \midrule
9 & I can tell that the people of Sauri must of thought that children needed education money not so they didn't ask people for the school fees, and the kids wouldn't go hungry during school hours they served the children lunch. &  & Delete & Content & LCE Reasoning & Successful \\ \midrule
10 & ... & ... & ... & ... & ... & ... \\  \bottomrule
\end{tabular}
\end{adjustbox}
\caption{Example of revision annotations for an \textit{elementary essay}. Note that the successful and unsuccessful labels in the last column are only used for evidence and reasoning content revisions; other purpose types in \cite{zhang2017corpus} are not in the scope of this study as we only focus on evidence use and reasoning in argumentative writing.}
\label{table:data-example-elementary}
\end{table*}

\subsection{Data Collection}
AR corpora are  rarely annotated in the revision community because of their expensive annotation costs. The publicly available~\textit{college essay} corpus for AR quality predictions~\cite{afrin2023predicting} contains paired drafts of argumentative essays written in response to an essay prompt (original drafts) and revised based on feedback (revised drafts). The corpus is comprised of 60 essays (N=60 college students), inclusive of both native and proficient non-native speakers of English, in response to an essay prompt about Technology Proliferation.  Students received general feedback upon completion of their first drafts, asking them to add more examples in their second drafts. The second drafts then received non-textual feedback through the ArgRewrite system~\cite{zhang2016argrewrite} to help students write their third drafts. Afterward, the second and third drafts were collected as pairs of original and revised drafts. 

We followed a similar protocol to collect 596 \textit{elementary essays} written by grade 5 to 6 students who were taking the Response to Text Assessment~\cite{correnti2013assessing}. 296 students wrote an essay in response to a prompt about the United Nation's Millenium Villages Project (MVP). The students then revised their essays in response to formative feedback from an Automatic Writing Evaluation (AWE) system that used rubric-based algorithms to assess the quality of evidence use and reasoning~\cite{zhang2019erevise,wang2020eRevise}. The other 300 students did the same tasks for an essay prompt about Space Exploration (Space). We combined the collected essays from the two essay prompts because students shared similar argumentative writing skills and the scoring rubric and feedback messages were constant across prompts.

\subsection{Preprocessing}
We preprocessed collected \textit{elementary essays} for annotations. First, sentences from original and revised drafts were aligned into pairs of original sentence (OS) and revised sentence (RS) using a sentence alignment tool Bertalign~\cite{liu2022bertalign}. The aligned pairs were programmingly labeled with \textit{no change} if OS and RS are the same, \textit{modifying} if OS and RS are not empty but not same, \textit{adding} if OS is empty but RS not, or \textit{deleting} if RS is empty but OS not. The changed alignments were automatically classified into \textit{surface} and \textit{content} revisions by a pretrained classifier. Note that  the sentence alignments and classification were first done by the system and then manually justified and corrected by annotators, and only aligned \textit{content} revisions were used for annotations. 

\subsection{Annotations}
We used the Revisions of Evidence use and Reasoning (RER) scheme~\cite{afrin2020RER} to annotate revision purposes in \textit{elementary essays}, which encodes the nature of students’ revision of evidence use and reasoning. Evidence use refers to the selection of relevant evidence from a given source article to support a claim, while reasoning means a reasoning process of connecting the evidence to the claim. Thus, the \textit{content} revisions are annotated with claim-related, evidence, and reasoning revisions. The RER scheme only applies to evidence and reasoning, where evidence revisions were labeled with \textit{relevant, irrelevant, repeated evidence, non-text based} and \textit{minimal}, and reasoning revisions were labeled with \textit{linked claim-evidence (LCE), not LCE, paraphrase evidence, generic, commentary, and minimal}.

Furthermore, we followed the AR quality scheme~\cite{afrin2023predicting} to programmingly encode annotated RER labels (revision purposes) into \textit{successful} and \textit{unsuccessful} revisions. The \textit{relevant} evidence was encoded as \textit{successful} while the \textit{repeated, non-text based}, and \textit{minimal} evidence were encoded as \textit{unsuccessful}. The \textit{LCE} and \textit{paraphrase} reasoning were encoded as \textit{successful}. The \textit{not LCE, paraphrase evidence, generic, commentary}, and \textit{minimal} reasoning were encoded as \textit{unsuccessful}. Table \ref{table:sentence-stats} shows label distributions in \textit{elementary essays} and \textit{college essays} where \textit{elementary essays} have almost an even number of reasoning and evidence annotations. The \textit{adding} revisions are the most frequent ARs across two essays. Samples of annotations for \textit{elementary essays} and \textit{college essays} are shown in Table \ref{table:data-example-elementary} and Table \ref{table:data-example-college} (in Appendix \ref{sec:appendix}), respectively. In practice, the RER annotations were done by one expert annotator. We sampled about 20 percent of annotated essays about both Space and MVP prompts and asked another well-trained annotator to annotate the sampled essays. The two-annotator Kappa scores are shown in Table~\ref{table:annotation-agreement}.

\begin{table}[t]\small
\centering
\begin{adjustbox}{width=0.89\columnwidth}
\begin{tabular}{ccccc}
\toprule
          & \multicolumn{2}{c}{\textbf{Space Essays}} & \multicolumn{2}{c}{\textbf{MVP Essays}} \\ \cmidrule{2-5} 
          &  RER\#            & Kappa            & RER\#            & Kappa           \\ \midrule
Reasoning & 148                 & 0.86             & 135                 & 0.84            \\
Evidence  & 108                 & 0.89             & 136                 & 0.80            \\ \bottomrule
\end{tabular}
\end{adjustbox}
\caption{Annotation agreement for reasoning and evidence RER annotations in a sample of 20 percent of \textit{elementary essays} regarding Space and MVP prompts.}
\label{table:annotation-agreement}
\end{table}

\section{Methods}
\subsection{Preliminary}
\label{sec:preliminary}
In this section, we introduce notations for the AR quality prediction task. We denote $R_1$ and $R_2$ as original and revised sentences in the original and revised drafts, respectively. In particular, $R_1$ is always empty in \textit{adding} ARs (e.g., row \#6 in Table \ref{table:data-example-elementary}); $R_2$ is always empty in \textit{deleting} ARs (e.g., row \#9 in Table \ref{table:data-example-elementary}); neither $R_1$ nor $R_2$ are empty in \textit{modifying} ARs (e.g., row \#1 in Table \ref{table:data-example-elementary}). Thus, we only use $R_1$ in \textit{deleting} and $R_2$ in \textit{adding} ARs. In terms of \textit{modifying} ARs, we only use $R_2$ because $R_2$ is a revised version of $R_1$ thus are very close to $R_1$ (e.g., row \#2 in Table \ref{table:data-example-elementary}). In addition, we denote ACs as a couple of sentences related to ARs in their corresponding drafts, where $C_1$ represents the ACs of $R_1$ in the original draft and $C_2$ represents the ACs of $R_2$ in the revised draft (details in Sec. \ref{sec:context}), respectively. To this end, we formulate the task of predicting AR quality as classifying the AR-AC pairs $\{R_i, C_i\}$ into \textit{successful} and \textit{unsuccessful} labels, where $i=1,2$. Specifically, we use pair $\{R_1, C_1\}$ for \textit{deleting} and $\{R_2, C_2\}$ for \textit{adding} and \textit{modifying} ARs.

\subsection{Argumentative Context}
\label{sec:context}

Consistent with long-established models of argumentation such as Toulmin’s model~\cite{toulmin1958uses}, well-developed arguments are characterized by the alignment of claim, evidence, and warrants (i.e., reasoning related to why the evidence supports the claim)~\cite{reznitskaya2008learning}. For example, the appropriateness of a piece of evidence for advancing an argument is context-dependent because that judgment is determined relative to an author's prior claim(s) or reason(s). As a case in point, the unsuccessful AR \#372 shown in Figure~\ref{fig:example} would have been unobservable absent an understanding of the author’s claim or argument’s context. Recent work by~\citet{afrin2023predicting} has used short and long text segments immediately before and after the AR as context for predicting AR quality, however, the study has some significant drawbacks. First, the window size of the contexts is an unpredictable parameter because a reasoning sentence could refer to the evidence far ahead of the AR (e.g., reasoning in row \#6 refers to the evidence in row \#1 in Table~\ref{table:data-example-elementary}). Second, location-based contexts did not explain why ACs make a difference to ARs from an argumentative perspective and thus fail to analyze the argumentative roles of ACs in AR quality predictions. As we noted above, the evaluation of a reasoning sentence as desirable depends on whether it appropriately references evidence or claims in the student's essay, but this relationship has not been explored in prior revision research. Thus, in the current study, we define three ACs to study their relationship to AR quality: \textbf{(1) AC-Claim}: the context containing essay claims or arguments; \textbf{(2) AC-Reasoning}: the context containing reasoning related to the claim or evidence in the essay; \textbf{(3) AC-Evidence}: the context containing evidence to support or oppose claims.

\subsection{ChatGPT Prompts}
Pretrained ChatGPT on a series of GPT3.5 models has shown promising results in solving information extraction~\cite{Li2023EvaluatingCI} and summarization~\cite{Yang2023ExploringTL} tasks in zero-shot settings, however, doing the two tasks at the same time has not been explored in generating ACs. Therefore, we developed two versions of ChatGPT prompts that generate useful ACs for predicting AR quality: (1) Single prompts that generate ACs in one pass and (2) Chain-of-Thought prompts that generate ACs in two passes. 

\subsubsection{Single Prompts}
In this section, we introduce Single prompts for AC generations. Basically, we need ChatGPT to generate useful ACs for AR quality predictions by reading the student essays. We limit the generation to a two-sentence length for two reasons. First, the generated ACs will be used in an AR-AC pair $\{R_i, C_i\}$, where $R_i$ is normally one sentence, thus long ACs ($C_i$) paired with short ARs ($R_i$) will make the AR quality prediction model (introduced in Sec.~\ref{sec:desirability-model}) learn to attend to the context rather than the revisions. Second, the most intuitive location-based baseline (Base-Short in Sec.~\ref{sec:experiment}) uses the adjacent sentences before and after target ARs, which contain at most two sentences. Therefore we limit the generations to exact two sentences, which can be done with a single zero-shot prompt \texttt{please summarize [X] in the essay [Y] in two sentences}, where \texttt{[X]} slot is one of the \textit{claim, reasoning}, and \textit{evidence}, \texttt{[Y]} is an input essay. 

\subsubsection{Chain-of-Thought Prompts}
In addition to Single prompts, Chain-of-Thought (CoT) prompts~\cite{cot_wei} are conceptually simple yet effective in multiple reasoning tasks. We adopt this idea and use zero-shot-CoT prompts to generate ACs, which run prompting in two passes but do not require step-by-step few-shot examples.

\textbf{The first-pass CoT prompt to extract ACs.} The first pass of the CoT prompts is to extract claim, evidence, and reasoning sentences from input essays. We aim to extract exact sentences from input essays without introducing any external knowledge in ChatGPT itself. The first-pass prompt is: \texttt{please list [X] sentences in the essay [Y]}, where \texttt{[X]} slot is chosen from one of the \textit{claim, reasoning}, and \textit{evidence}, and \texttt{[Y]} is an input essay. The extracted ACs are formulated as a list of sentences from the input essays, where the length of the list ranges from one to dozens because \textit{claim, reasoning}, and \textit{evidence} have multiple instances in an essay. To make sure the extracted ACs are informative and not exceeding the word limits of DistilRoBERTa encoders (see Sec. \ref{sec:desirability-model}), we perform summarization on the extracted sentences in the second-pass prompt.

\begin{figure*}[t]
    \centering
    \includegraphics[width=2.1\columnwidth]{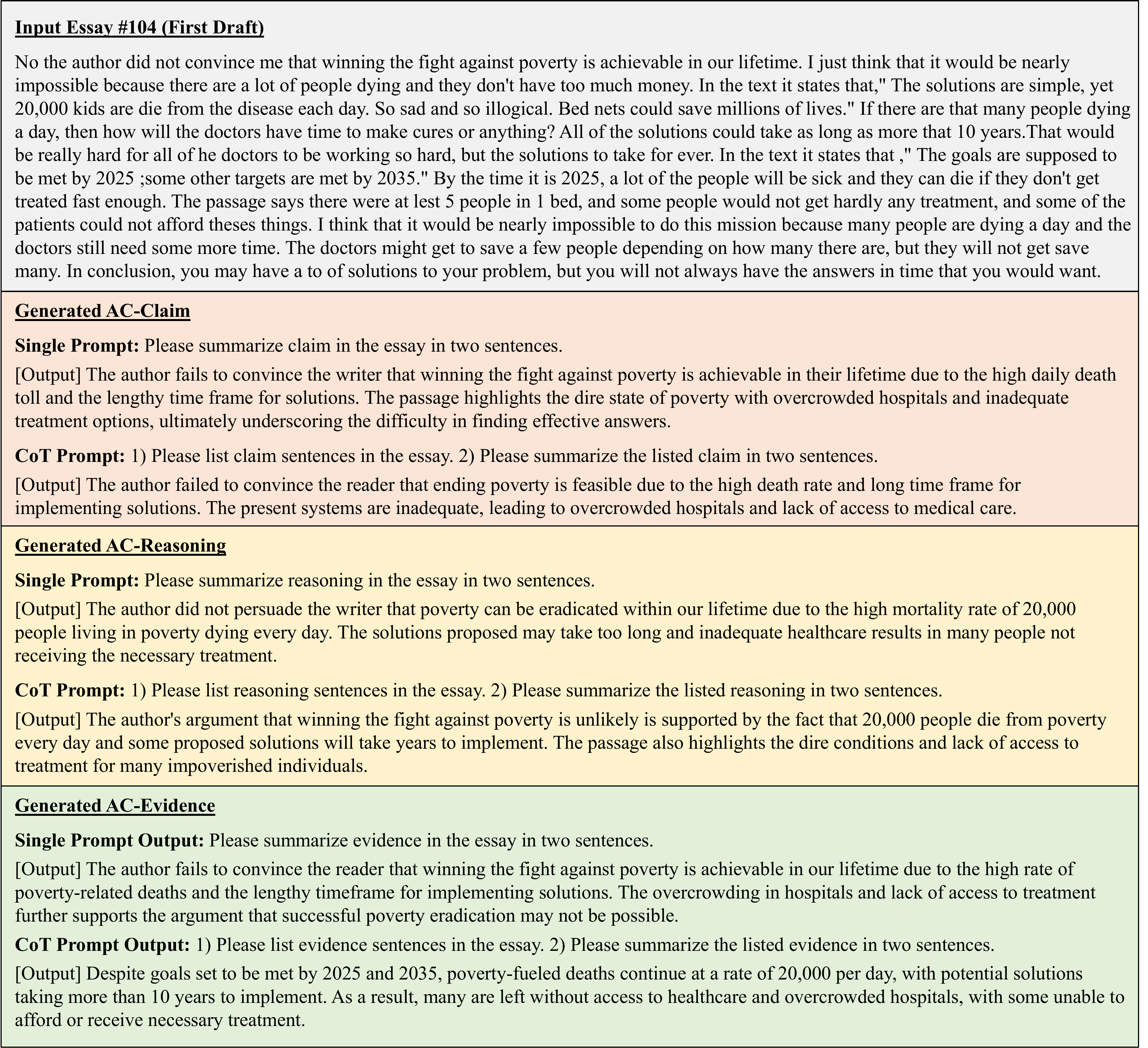}
    \caption{The input and output of the ChatGPT with zero-shot Single and CoT prompts for an \textit{elementary essay}.}
    \label{fig:example-second-prompt-elementary}
\end{figure*}

\textbf{The second-pass CoT prompt to summarize ACs.} The second prompt is continued to the first prompt, following an extraction-summarization CoT. The prompt is \texttt{please summarize [X] in two sentences}, where \texttt{[X]} slot is chosen from the claim, reasoning, and evidence sentences extracted in the first prompting pass, which ensures the outputs in a length of exact two sentences. Figure \ref{fig:example-second-prompt-elementary} and Figure~\ref{fig:example-second-prompt-college} (in Appendix~\ref{sec:appendix}) show examples of the zero-shot Single and CoT prompts that help ChatGPT generate ACs in \textit{elementary} and \textit{college essays}, respectively.

\subsection{AR Quality Prediction}
\label{sec:desirability-model}
\begin{figure}[t]
    \centering
    \includegraphics[width=0.83\columnwidth]{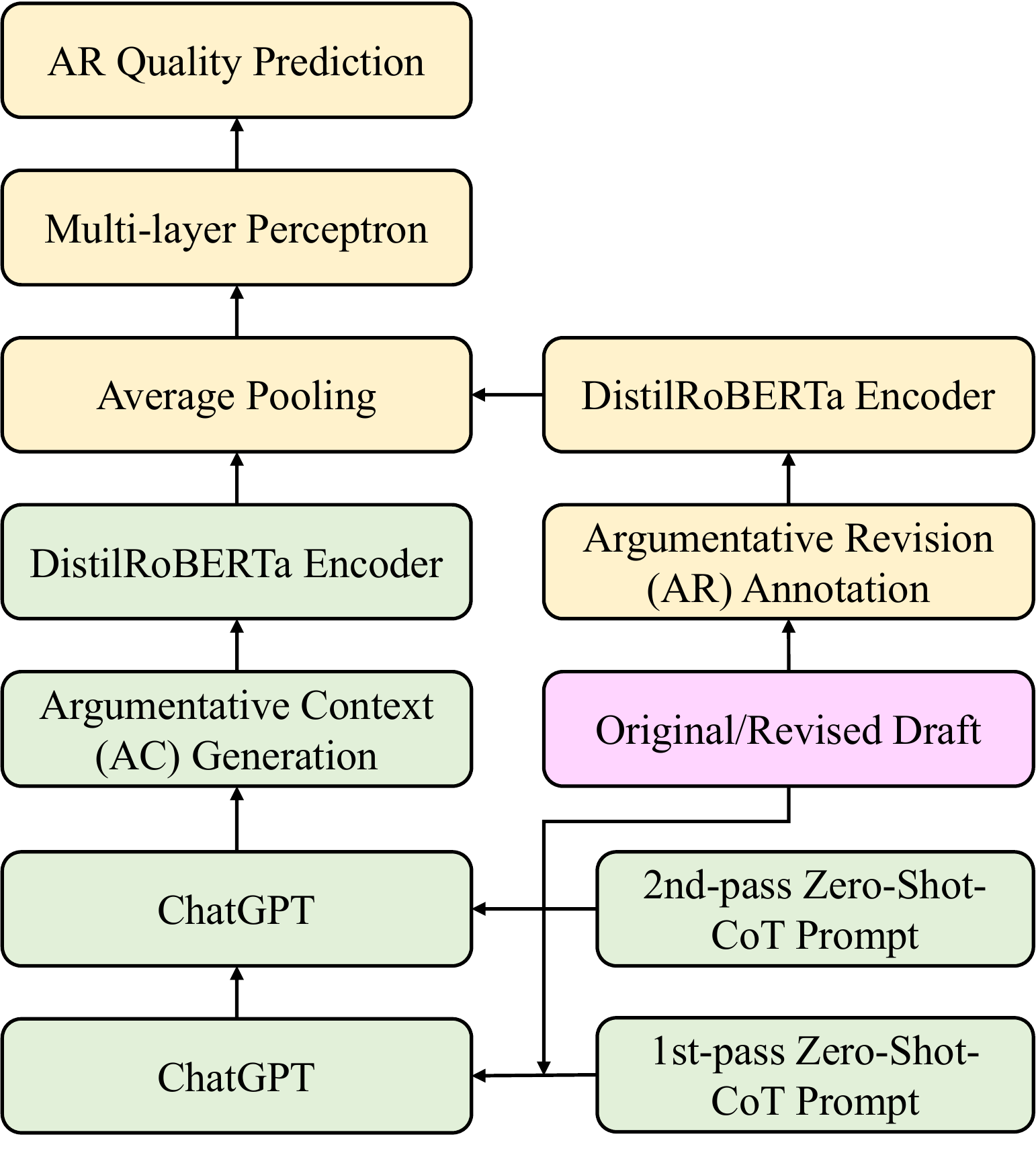}
    \caption{The overall framework of AR quality predictions, where the pink box is input; the green boxes are our proposed; the yellow boxes are existing methods.}
    \label{fig:framework}
\end{figure}

We define AR quality prediction as a binary classification of AR-AC pair $\{R_i,C_i\}$ (see Sec.~\ref{sec:preliminary}), where $R_i$ is annotated and $C_i$ is generated by ChatGPT. Prior works employed BERT-BiLSTM architecture to train revision classifiers~\cite{anthonio-roth-2020-learn,afrin2023predicting}. Instead, we use DistilRoBERTa~\cite{Sanh2019DistilBERTAD} as text encoders for both annotated ARs and ChatGPT-generated ACs. The last hidden layers of the DistilRoBERTa encoders are fed to an average-pooling layer and then connected to a multi-layer perception classifier that contains a sequence of batch normalization layer, ReLU layer, dropout layer with a 0.5 rate, and Sigmoid layer. The overall framework is shown in Figure \ref{fig:framework}.

\section{Experiments}
\label{sec:experiment}

\begin{table*}[t]\small
\centering
\begin{adjustbox}{width=\textwidth}
\begin{tabular}{lc|ccc|ccc|ccc}
\toprule
 \multirow{2}{*}{Contexts} & \multirow{2}{*}{Prompts} & \multicolumn{3}{c|}{\textbf{Resoning \& Evidence ARs}} & \multicolumn{3}{c|}{\textbf{Reasoning ARs}} & \multicolumn{3}{c}{\textbf{Evidence ARs}} \\
 & & Precision & Recall & F1 & Precision & Recall & F1 & Precision & Recall & F1 \\ \midrule
Base-Short & N/A    & 67.79 & 67.17 & 67.29 & 70.29 & 70.01 & 69.94 & 63.42 & 62.60 & 62.67 \\
Base-Long     & N/A    & 68.45 & 68.01 & 68.06 & 69.99 & 69.76 & 69.71 & 65.38 & 64.69 & 64.71 \\ \midrule
AC-Claim      & Single & \underline{70.31} & \underline{69.83} & \underline{69.91} & \underline{72.63} & \underline{72.47} & \underline{72.38} & 66.60 & 65.33 & 65.57 \\
AC-Reasoning  & Single & \underline{70.15} & \underline{69.67} & \underline{69.74} & 72.10 & \underline{71.95} & \underline{71.83} & 66.57 & 65.60 & 65.79 \\
AC-Evidence   & Single & \underline{70.28} & \underline{69.97} & \underline{69.93} & \underline{72.46} & \underline{72.31} & \underline{72.13} & 66.55 & 65.87 & 65.85 \\ \midrule
AC-Claim      & CoT    & \underline{70.09} & \underline{69.64} & \underline{69.74} & \underline{71.83} & \underline{71.76} & \underline{71.70} & \,\,66.74* & \,\,65.71* & \,\,65.88* \\
AC-Reasoning  & CoT    & \,\, \underline{\textbf{71.14}}* & \,\,\underline{\textbf{70.81}}* & \,\,\underline{\textbf{70.81}}* & \,\,\underline{\textbf{72.86}}* & \,\,\underline{\textbf{72.80}}* & \,\,\underline{\textbf{72.63}}* & \,\,\underline{\textbf{68.00}}* & \,\,\underline{\textbf{67.00}}* & \,\,\underline{\textbf{67.16}}* \\
AC-Evidence   & CoT    & \,\,\,\underline{70.43}* & \,\,\underline{70.03}* & \,\,\underline{70.01}* & \,\,\underline{72.48}* & \,\,\underline{72.34}* & \,\,\underline{72.20}* & \,\,66.76* & \,\,66.06* & \,\,66.05* \\ \bottomrule
\end{tabular}
\end{adjustbox}

\caption{Experimental results on \textit{elementary essay} corpus. The bold numbers are the best results. The underlined numbers statistically outperformed the strong (Base-Long) baseline in a paired t-test with $p<0.05$. The asterisks indicate zero-shot-CoT prompts are better than zero-shot-Single prompts.}
\label{table:main-results}
\end{table*}

To answer the RQs, we implemented location-based contexts as baselines, and a series of ACs as comparable methods:
\begin{itemize}[leftmargin=*,itemsep=-2pt, topsep=3pt]
  \item \textbf{Base-Short}: We implement a standard revision prediction baseline that uses the adjacent sentences immediately before and after a revision as contexts~\cite{afrin2023predicting}. 
  \item \textbf{Base-Long}: We implement a strong baseline that considers all the sentences that are revised around a target revision until an unchanged sentence is found~\cite{afrin2023predicting}.
  \item \textbf{AC-Claim}: We use AC-Claim as the contexts that are generated by zero-shot Single and CoT prompts, respectively.
  \item \textbf{AC-Reasoning}: We use AC-Reasoning as contexts. The two versions use zero-shot Single and CoT prompts, respectively.
  \item \textbf{AC-Evidence}: We use AC-Evidence as contexts. The two versions are generated by zero-shot Single and CoT prompts, respectively.
\end{itemize} 

In the implementation, we built the framework pipeline with PyTorch\footnote{\url{https://pytorch.org}} and generated two versions of ACs using ChatGPT3.5-turbo API\footnote{\url{https://platform.openai.com}}. We used pretrained DistilRoBERTa-Base from Huggingface\footnote{\url{https://huggingface.co}} as text encoders, and optimized cross-entropy loss with Adam optimizer on a GeForce RTX 3090 GPU. We set the batch size as 16 and the learning rate as 5e-5 with 5\% decays every 4 epochs. We conducted 10-fold cross-validation, where 80\% of each 9-fold set was used for training, 20\% for parameter tuning, and the rest 1-fold set for testing. Finally, we ran the ChatGPT generation and the experiment pipeline three times and reported 3-seed-average macro Precision, Recall, and F1 on all the test sets. The implementation code is available at~\url{https://github.com/ZhexiongLiu/Revision-Quality-Prediction}.

\begin{table}[t]\small
\centering
\begin{adjustbox}{width=1\columnwidth}
\begin{tabular}{lc|ccc}
\toprule
Contexts & Prompts & Precision & Recall & F1 \\ \midrule
Base-Short & N/A & 59.95 & 60.06 & 58.61 \\
Base-Long & N/A & 61.21 & 61.60 & 59.48 \\ \midrule

AC-Claim & Single & 63.50 & 64.08 & 62.27 \\
AC-Reasoning & Single & 63.06 & 63.62 & 61.86 \\
AC-Evidence & Single & \underline{65.76} & 66.40 & \underline{64.71} \\ \midrule
AC-Claim & CoT & \,\,64.01* & \,\,64.96* & \,\,62.93* \\
AC-Reasoning & CoT & \,\,63.84* & \,\,64.33* & \,\,62.74* \\
AC-Evidence & CoT & \,\,\underline{\textbf{68.20}}* & \,\,\underline{\textbf{68.05}}* & \,\,\underline{\textbf{66.32}}* \\ \bottomrule
\end{tabular}
\end{adjustbox}
\caption{Experimental results on reasoning ARs in \textit{college essays}. The bold numbers are the best results. The underlined numbers statistically outperformed the strong (Base-Long) baseline in a paired t-test with $p<0.05$. The asterisks indicate zero-shot-CoT prompts are better than zero-shot-Single prompts.}
\label{table:main-college}
\end{table}

\section{Results and Discussion}
Table \ref{table:main-results} shows the experimental results for different sets of revisions from the \textit{elementary essay} corpus: all  reasoning and evidence revisions, just reasoning revisions, and just evidence revisions, respectively. We observed that both the proposed Single and CoT versions of ACs outperformed both baselines, with many of the CoT ACs  significantly better than the strong (Base-Long) baseline. This answered \textbf{RQ1} that ACs can help AR quality predictions. In reasoning ARs, excellent performance was yielded in using AC-Claim, AC-Reasoning, and AC-Evidence. This is because reasoning ARs might need claims to verify their usefulness and incorporate evidence and reasoning to check their relevance. Moreover, evidence ARs achieved the best with AC-Reasoning, which makes sense that identifying evidence AR requires related reasoning contexts that have information linking the evidence. Another interesting finding is that the Base-Long performed better than the Base-Short in evidence ARs but worse in reasoning ARs. This suggests that the longer context is not always helpful in the case that evidence contexts are usually sparsely distributed in the essay so the longer context will introduce more noise. It also suggests that reasoning sentences are mostly adjacent to other reasoning contexts and can be well captured by neighboring sentences. Furthermore, the observation that reasoning ARs results are generally better than evidence ARs indicates that reasoning ARs might be self-justifiable which means it might require fewer contexts than the evidence to identify AR quality. These observations answered \textbf{RQ2} that reasoning contexts are mostly useful, and both reasoning, claim, and evidence contexts benefit AR quality predictions. In addition, CoT prompts are generally better than Single prompts in most reasoning and evidence ARs, which indicates that identifying AR quality requires some contexts that might not be generated with Single prompts. This answered \textbf{RQ3} that CoT prompts are generally better than Single prompts.

We also evaluated the effectiveness of ACs on the \textit{college essay} benchmark. Note that~\citet{afrin2023predicting} conducted data augmentation with a simple synonym replacement because they argued that it was impossible to obtain reasonable results without training on augmented data. We hypothesized that data augmentation will introduce noise but the limited data can yield reasonable results training with the DistilRoBERTa-based model. Therefore, we did not do data augmentation and compared AC-based methods to our implemented standard and strong baselines on reasoning revisions (excluding the rare evidence revisions as shown in Table~\ref{table:sentence-stats}). Results in Table~\ref{table:main-college} show that the DistilRoBERTa model is able to learn from even small-size data without data augmentation. In addition, AC-based methods perform better than both the standard and strong baselines, where AC-Evidence has significant improvement. This again suggests that ACs are generally useful for predicting AR quality and CoT prompts are generally better than Single prompts for generating useful ACs. Moreover, we observed that AC-Evidence generated by Single and CoT prompts is better than the other ACs. It is slightly different from the reasoning column in Table~\ref{table:main-results}. This might suggest that revisions in \textit{college essays} may focus on evidence revisions that match generated evidence ACs. However, \textit{claim} and \textit{reasoning} results have similar F1 scores across two versions of prompts, which might suggest the extracted AC-Claim and AC-Reasoning are similar in \textit{college essays} (e.g., prompting outputs in Figure \ref{fig:example-second-prompt-college} in Appendix~\ref{sec:appendix}), which might be because \textit{college essays} have claim and reasoning sentences disentangled. In general, CoT prompts are somewhat better than Single prompts in AC-Claim and AC-Reasoning generation, and both Single and CoT prompts are promising in AC-Evidence generation.

\begin{figure}[t]
    \centering
    \includegraphics[width=0.98\columnwidth]{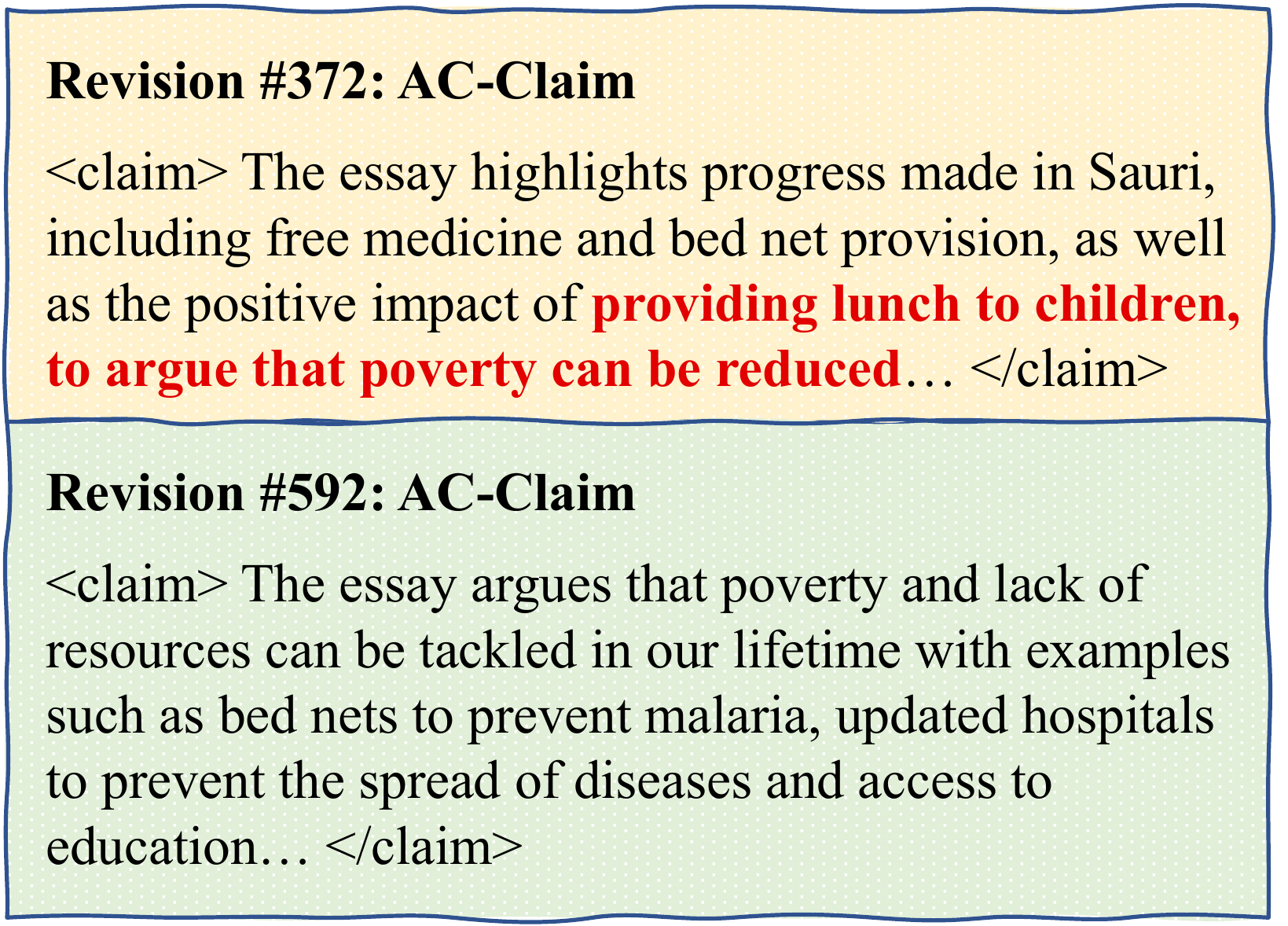}
    \caption{Two pieces of ChatGPT-generated AC-Claims. The red bold is the context to identify Revision \#372 is a \textit{already existed} adding, while \#592 is a \textit{relevant} adding toward their contexts in Figure \ref{fig:example}.}
    \label{fig:example-correct-context}
\end{figure}

As a case study, we examine the effectiveness of ACs in Revision \#372 and \#592 presented in Figure~\ref{fig:example}. The ChatGPT-generated AC-Claim is shown in Figure \ref{fig:example-correct-context}, where the red bold sentence \textit{``providing lunch to children, to argue that poverty can be reduced''} is helpful to identify that the added sentence, \textit{``It was hard for them to concentrate though, as there was no midday meal.''} in Revision \#372 is a \textit{already existed} evidence, and thus it was an \textit{unsuccessful} revision. However, AC-Claim in Revision \#592 does not show repeated but \textit{relevant} information, and thus the AR is regarded as \textit{successful}.

\section{Conclusion} 
This work studies the relationship between Argumentative Contexts (ACs) and Argumentative Revisions (ARs) in argumentative writing. In particular, we use zero-shot-CoT prompts to facilitate ChatGPT-generated ACs for AR quality predictions. The experiments on our  \textit{elementary essays} corpus and publicly available \textit{college essays} benchmark demonstrate the superiority of the proposed ACs over existing location-based context baselines, which proposes a new direction for predicting AR quality. The analysis suggests that most evidence ARs need reasoning ACs, and reasoning ARs need a diverse set of claims, evidence, and reasoning ACs to predict their quality.

\section{Limitations}
Our experiments were built on perfect sentence alignments in the original and revised essay drafts, thus the performance could be lower in the real end-to-end Automated Writing Evaluation (AWE) system. In addition, our corpus is small due to expensive annotation processes, which makes it challenging to train or finetune large language models. Also, we only focus on revisions in argumentative writing, specifically, we focus on the evidence and reasoning revisions, however other revisions like claim revisions are not used. Furthermore, the revised drafts were done after providing feedback on the original drafts, which means the revised student essays are likely to follow the instructions in the feedback but we did not use this information for revision quality predictions, which will be used in our future work.

Our proposed Argumentative Contexts (ACs) are generated by ChatGPT which is not free for the whole community. Also, ChatGPT-generated ACs have small randomness, which is also the reason we did 3-seed runs in the experiments. In addition, the ACs are essay-level context which means different revisions in the same essay use the same context. It could be tailored to have sentence-level ACs where each sentence-level revision has slightly different revision purposes, but it would cost more time and money. Moreover, our proposed zero-shot-CoT prompts perform better than Single prompts by small margins in specific cases, which indicates that Chat-GPT is limited to conducting CoT extraction and summarization to handle complex wording and sentence structure. Therefore, we might need to redesign the prompts in our future work.

\section{Ethics}
Our corpus was collected under standard protocols that were approved by an institutional review board. Our annotated data is not publicly available which ensures the safety of private information of the students, and thus will not pose any ethical concerns because other researchers can not access our data and replicate our results. Our future work is to incorporate proposed methods in real AWE systems to evaluate student writings and provide informative feedback based on  predictions. But there is a risk that the system might give poor advice based on incorrect AR quality predictions, given that the model may learn biases with small annotated data.

\section*{Acknowledgments}
The research was supported by the National Science Foundation Award \#2202347 and a gift from CloudBank. The opinions expressed were those of the authors and did not represent the views of the institutes. We would like to thank anonymous reviewers and Pitt PETAL group for their valuable feedback on this work.

\bibliography{anthology,custom}
\bibliographystyle{acl_natbib}

\appendix
\section{Appendix}
\label{sec:appendix}

\begin{table*}[!htb]\small
\centering
\begin{tabular}
{p{0.01\textwidth}p{0.26\textwidth}p{0.26\textwidth}p{0.05\textwidth}p{0.06\textwidth}p{0.08\textwidth}p{0.09\textwidth}}		
\toprule
\textbf{ID} & \textbf{Original Draft Sentence} & \textbf{Revised Draft Sentence} & \textbf{Revision} & \textbf{Revision Type} & \textbf{Revision Purpose} & \textbf{Quality Label} \\ \hline
1 & A mother who would have no other   way of reaching her children can easily speak to them or leave a message via   voicemail. & A mother who would have no other   way of reaching her children can easily speak to them or leave a message via   voicemail. & N/A & N/A & N/A & N/A \\ \hline
2 & Technology makes it possible to   reach anyone at any time. &  & Delete & Content & LCE Reasoning & Unsuccessful \\ \hline
3 &  & In addition, technology makes it   possible to increase the amount of communication between people drastically. & Add & Content & LCE Reasoning & Successful \\ \hline
4 & … & … & ... & … & … & … \\ \hline
5 & People from different continents   who may have never met before can now have conversations every day; even   those from a remote location are available to the world, provided they have   the Internet. & People from different continents   who may have never met before can now have conversations every day; even   those from a remote location are available to the world, provided they have   both the Internet and a corresponding device. & Modify & Surface & N/A & N/A \\ \hline
6 &  & How could a cold inanimate   screen replace seeing the emotions and expressions of a loved one? & Add & Content & not LCE Reasoning & Unsuccessful \\ \hline
7 &  & An essential thing to consider   is that while perhaps it may be harder to convey one's full message complete   with feelings through the Internet, the fact remains that in a changing world   where people are busier and farther away, electronic devices are helping   everyone keep in contact with each other at any time of the day and at any   location. & Add & Content & LCE Reasoning & Successful \\ \hline
8 &  & Those who argue for the   retardation of technology simply cannot accept that the world is developing   to be more tech driven; as more and more people have electronic devices, they   are also changing to be more used to this kind of communication. & Add & Content & not LCE Reasoning & Unsuccessful \\ \hline
9 & … & … & ... & … & … & … \\ \hline

\end{tabular}
\caption{Example of revision annotations for a \textit{college essay}.}
\label{table:data-example-college}
\end{table*}

\begin{figure*}[t]
    \centering
    \includegraphics[width=2.1\columnwidth]{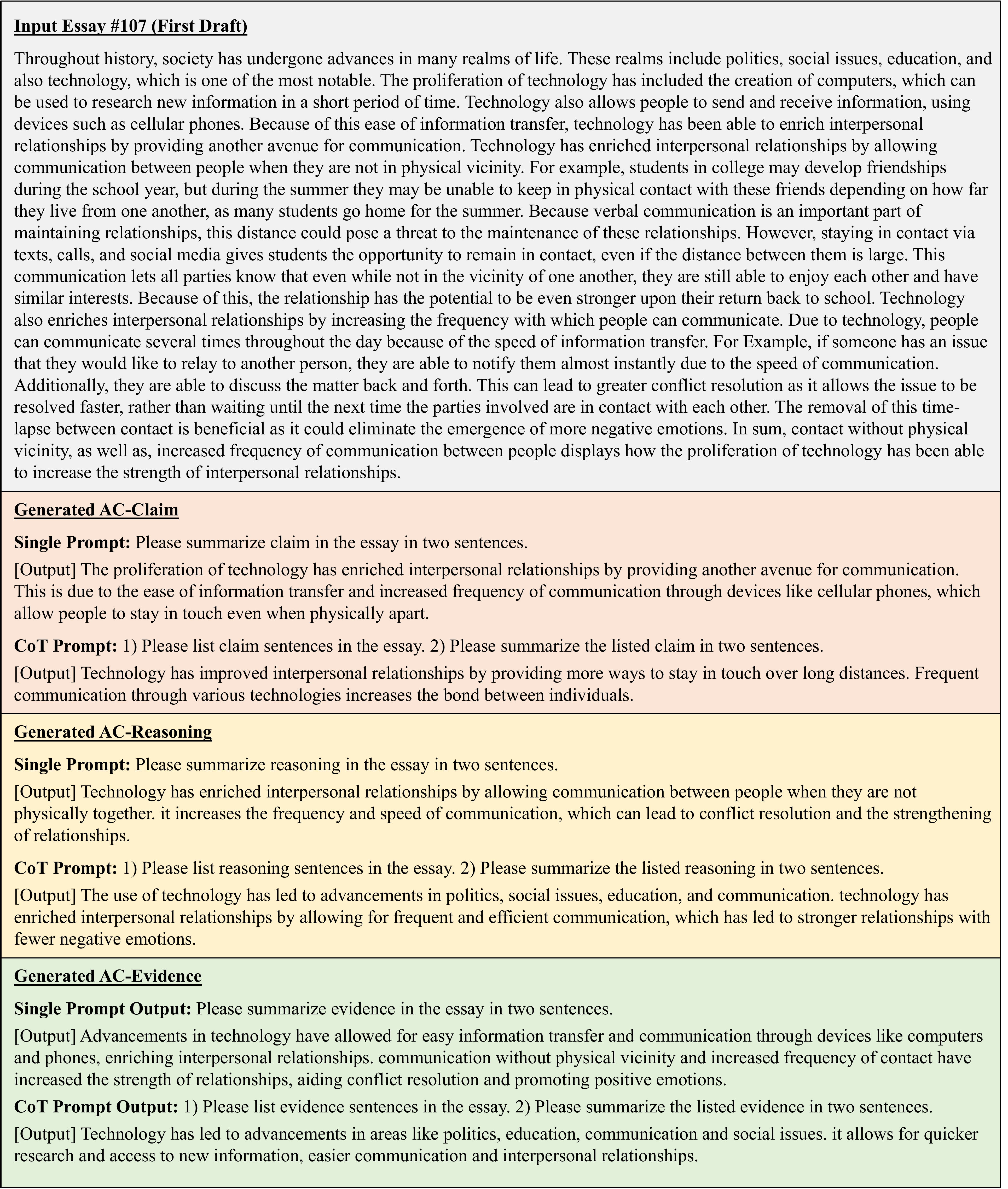}
    \caption{The input and output of the ChatGPT with zero-shot Single and CoT prompts for a \textit{college essay}.}
    \label{fig:example-second-prompt-college}
\end{figure*}
\end{document}